\documentclass[runningheads]{llncs}

\usepackage{times}
\usepackage{soul}
\usepackage{url}
\usepackage{hyperref}

\usepackage[utf8]{inputenc}
\usepackage[small]{caption}
\usepackage{graphicx}
\usepackage{amsmath}
\usepackage{booktabs}
\usepackage{algorithm}
\usepackage{arevmath}    
\usepackage[noend]{algpseudocode}
\usepackage{listings,multicol}
\usepackage{multicol}

\usepackage[caption=false]{subfig}

\usepackage{oz}
\usepackage{makeidx}
\usepackage{listings}
\usepackage{amsfonts}
\usepackage{relsize}
\usepackage{urwchancal}
\usepackage{stmaryrd}
\usepackage{ dsfont }

\usepackage{float}
\newfloat{algorithm}{t}{lop}

\newcommand{\uri}[1]{\ensuremath{{:}{\texttt{#1}}}}

\usepackage[misc,geometry]{ifsym} 
\hypersetup{
     colorlinks = true,
     linkcolor = blue,
     anchorcolor = blue,
     citecolor = blue,
     filecolor = blue,
     urlcolor = blue
     }
     
\newcommand\blfootnote[1]{%
  \begingroup
  \renewcommand\thefootnote{}\footnote{#1}%
  \addtocounter{footnote}{-1}%
  \endgroup
}

\let\oldmaketitle\maketitle
\renewcommand{\maketitle}{\oldmaketitle\setcounter{footnote}{0}}

\begin{document}

\title{A Policy Editor for Semantic Sensor Networks}
\author{Paolo Pareti\ensuremath{^{(\textrm{\Letter})}} \inst{1} \and George	Konstantinidis \inst{1} \and Timothy J.\ Norman \inst{1}}

\authorrunning{P. Pareti et al.}

\institute{University of Southampton, Southampton, United Kingdom \\ \email{pp1v17@soton.ac.uk}}

\maketitle

\begin{abstract}

An important use of sensors and actuator networks is to comply with health and safety policies in hazardous environments. In order to deal with increasingly large and dynamic environments, and to quickly react to emergencies, tools are needed to simplify the process of translating high-level policies into executable queries and rules. We present a framework to produce such tools, which uses rules to aggregate low-level sensor data, described using the Semantic Sensor Network Ontology, into more useful and actionable abstractions. Using the schema of the underlying data sources as an input, we automatically generate abstractions which are relevant to the use case at hand. In this demonstration we present a policy editor tool and simulation on which policies can be tested.

\end{abstract}

\section{Introduction}

Within the area of Occupational Health and Safety (OHS) it is common to deploy sensor networks in hazardous working environments. The observations from these sensors are then processed by monitoring systems to ensure compliance with health and safety policies.  
high-level policies represent important principles, such as the need to ensure safe working environments, and the specification of which environmental factors should be considered as indications of unsafe conditions, such as a carbon monoxide (CO) concentration higher than $50 ppm$. These high-level policies are typically interpreted as more concrete and actionable policies, which are the subject of this work. 
\blfootnote{Copyright © 2019 for this paper by its authors. Use permitted under Creative Commons License Attribution 4.0 International (CC BY 4.0).}

For example, let us consider the policy: ``if the carbon monoxide  concentration of a tunnel exceeds $50 ppm$, personnel should be evacuated from that tunnel''. We can imagine how this policy, so far just described in natural language, can be fully automated by sensors and actuators. For instance, a monitoring system can trigger the evacuation alarm actuators in a tunnel as soon as sensors detect CO levels exceeding the limit.

At the moment, domain experts need to translate such policies into executable queries, however this process is slow, expensive and error prone. This is especially problematic when dealing with increasingly large and dynamic sources of data, as in the case of Internet of Things (IoT) applications. For example, the underlying sources of data might change as new sensors are deployed, or old ones malfunction. To quickly respond to changes, non-expert users require the ability to define and edit executable policies; that is, policies which can be translated into queries and be automatically monitored. In this work, we demonstrate an automatic approach to translate low-level sensor data, modelled according the Semantic Sensor Network Ontology (SSN)  \cite{Taylor2017SSN}, into higher level concepts. These concepts are the primitive constructs that users can use to create policies.
While retaining the ability to be directly translated into database queries, these higher level concepts hide the complexity of the underlying data models under natural language labels, which offer non-expert users a more intuitive way to work with sensor and actuators.

\section{Description of the Framework}

The main goal of this framework is to automatically aggregate sensor data into more useful abstractions. For example, let us consider mine tunnels fitted with carbon monoxide sensors. A SSN sensor reading can be represented by \texttt{RDF} triples matching the graph pattern in Fig. \ref{listing:basic}. In this example, the \texttt{URI}s \uri{CO} and \uri{Tunnel} denote, respectively, the concepts of carbon monoxide concentration, and mine tunnels. These four triple patterns describe: (1) that \texttt{?s} is an observation of carbon monoxide concentration, (2) that the value measured is \texttt{?b}, (3) that this measurement was done with respect to \texttt{?a} and (4) that \texttt{?a} is a tunnel.

\begin{figure}
\vspace{-6mm}

\footnotesize \begin{lstlisting}
?s sosa:observedProperty :CO .    ?s sosa:hasResult ?b .
?s sosa:hasFeatureOfInterest ?a . ?a rdf:type :Tunnel .
\end{lstlisting} \normalsize
\vspace{-5mm}
\caption{\label{listing:basic}}
\vspace{-6mm}
\end{figure}

Arguably, this type of data representation is hard to work with for non-experts, as it requires knowledge of \texttt{RDF} and SSN. In this work, we propose a more intuitive representation that relies on natural language and a basic understanding of variables (i.e.\ \texttt{?a} and \texttt{?b}). For example, Fig. \ref{listing:aggregate} shows such a representation of the \texttt{RDF} patterns in Fig. \ref{listing:basic}. In this representation variable \texttt{?s} is not explicitly mentioned.

\begin{figure}[h!]
\vspace{-6mm}
\begin{lstlisting}
"<@\textnormal{the carbon monoxide concentration of tunnel}@> ?a <@\textnormal{is}@> ?b"
\end{lstlisting} 
\vspace{-5mm}
\caption{\label{listing:aggregate}}
\vspace{-6mm}
\end{figure}

While the sentence in Fig. \ref{listing:aggregate} can be considered more suitable for human understanding, it does not clearify how it can be translated into an executable query. In order to combine the benefits of both representations, we create Abstract Concept Aggregations (ACA) which include both human-understandable labels, such as the one in Fig. \ref{listing:aggregate}, and their corresponding queriable representations, such as the graph pattern in Fig. \ref{listing:basic}.

\begin{figure}[!t]
     
         \center{\includegraphics[clip,width=0.99\columnwidth]{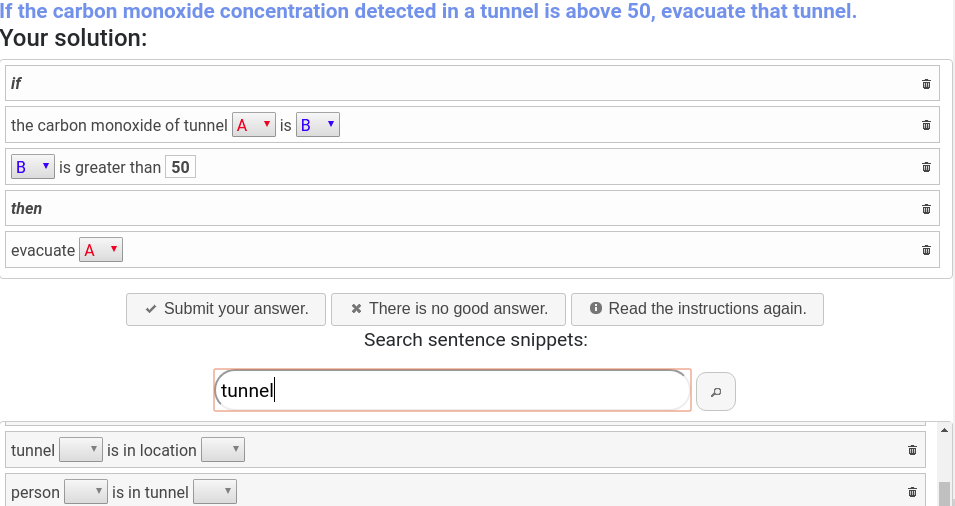}}
        
        \caption{\label{figure:Editor} Screenshot of the policy editor, with a sample formalisation of policy "If the carbon monoxide concetration in a tunnel is above 50, evacuate that tunnel.".}
        \vspace{-6mm}
\end{figure}

These ACA concepts enable us to create intuitive editor tools, such as the policy-editor tool displayed in Fig. \ref{figure:Editor}. This tool is designed to facilitate the composition of aggregate concepts and if-then rules by non-experts, hiding the complexity of the underlying data representations (in this case \texttt{RDF} and SSN). Using this tool, users can search for existing ACAs using keyword search, and then compose them together. Fig. \ref{figure:Editor} shows a natural language policy, and its corresponding formalisation using ACAs. It should be noted that this formalisation can be directly translated into a \texttt{SPARQL} query, and therefore it is immediately executable.

\begin{figure}[!t]
     
         \center{\includegraphics[clip,width=0.9\columnwidth]{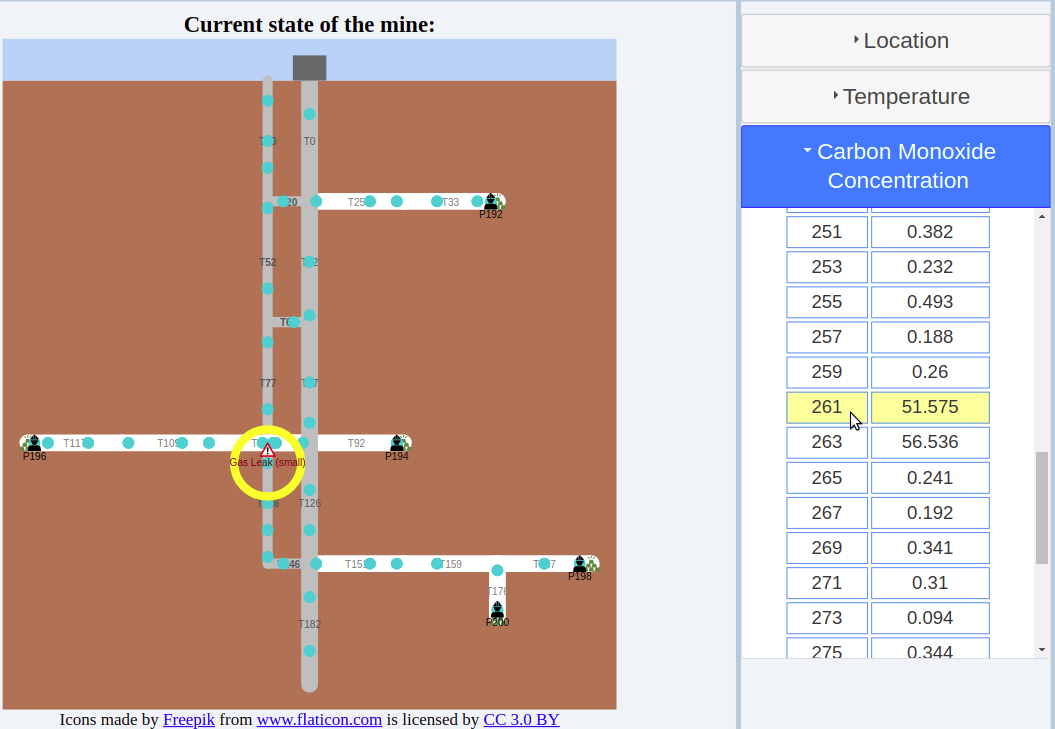}}
        
        \caption{\label{figure:MineSim} Screenshot of the mine simulator, showing the occurrence of a gas leak. This gas leak is detected by sensor 261, highlighted on the table on the right, and located in the centre of the yellow circle on the left, which shows an abnormally high concentration of Carbon Monoxide.}
\end{figure}

In order to test these policies, we have developed a simulation of a mining environment displayed in Fig. \ref{figure:MineSim}. This mining simulation captures simple, but common, features of real world mines, namely: (1) a layout of underground tunnels, (2) workers moving in the mine, whose location is captured by sensors they wear on their equipment, (3) environmental sensors, such as carbon monoxide and temperature sensors, (4) emergency situations, such as fire outbreaks and gas leaks and (5) several actuators, such as the possibility to evacuate the mine or geofence dangerous tunnel sections.

A core challenge of developing such a policy edtor lies in the generation of suitable ACAs. Manual construction of ACAs for each use case would require expert human intervention, and is therefore contrary to the main purpose of this framework. For this reason, our framework automatically generates ACAs using (1) the schema of the underlying sensor data sources and (2) a set of generic aggregation rules, defined a-priori for the application onotolgy (in this case, SSN). In order to construct new ACAs, aggregation rules do not simply infer new triples, but also specify how to construct their associated natural language labels. It should be noted that the rules used to generate the ACAs in our demonstration examples are not specific to the mining domain, and could be applied to any dataset that uses the SSN. For example, the rule that generates the label in Fig. \ref{listing:aggregate} from the schema of the triple patterns in Fig. \ref{listing:basic} is not aware of domain specific concepts such as carbon monoxide or tunnels.

In principle there is a large (possibly infinite) number of ACAs that can be constructed, and most of them might not be relevant. For example, an ACA aggregating observations from methane detectors is not relevant in an environment where methane detectors are not present. To tackle this, we reuse an existing approach for checking rule applicability on triplestores in different scenarios \cite{ParetiISWC2019}. We do this by extracting the notion of a triplestore schema for the particular scenario and devising an algorithm to check rule applicability against that schema. We use this method to check which aggregation rules are applicable in different scenarios/schemas and thus infer, and present to the user of our editor, those ACAs that are actually relevant.

\section{Demonstration}

In this demonstration we will present our policy editor (Fig. \ref{figure:Editor}) and mine simulator (Fig. \ref{figure:MineSim}). This demonstration aims to provide an intuitive understanding of our framework, its components, and its potential to simplify policy creation and editing by non-expert users. 
We will provide videos of the the different components of the framework in action. Sample videos of these systems and a link to an online demo can be found at this GitHub repository.\footnote{\url{https://github.com/paolo7/demo-files-ISWC2019}} 
Users will be able to directly interact with the system in order to (1) formalise policies using ACAs; (2) try modifications to the default set of aggregation rules and ontologies being used in order to evaluate the ACA generation; and (3) run the simulated mine environment and test the effectiveness of policies on it.

\bibliographystyle{splncs04}
\bibliography{litbib}

\end{document}